# Introspective Perception: Learning to Predict Failures in Vision Systems

Shreyansh Daftry, Sam Zeng, J. Andrew Bagnell and Martial Hebert

*Abstract*— As robots aspire for long-term autonomous operations in complex dynamic environments, the ability to reliably take mission-critical decisions in ambiguous situations becomes critical. This motivates the need to build systems that have situational awareness to assess how qualified they are at that moment to make a decision. We call this self-evaluating capability as introspection. In this paper, we take a small step in this direction and propose a generic framework for introspective behavior in perception systems. Our goal is to learn a model to reliably predict failures in a given system, with respect to a task, directly from input sensor data. We present this in the context of vision-based autonomous MAV flight in outdoor natural environments, and show that it effectively handles uncertain situations.

## I. INTRODUCTION

Perception is often the weak link in robotics systems. While considerable progress has been made in the computer vision community, the reliability of vision modules is often insufficient for long-term autonomous operation. Of course this is not surprising for two reasons. First, while vision algorithms are being systematically improved in particular by using common benchmark datasets such as KITTI [1] or PASCAL [2], their accuracy on these datasets does not necessarily translate to the real-world situations encountered by robotics systems. We advocate that for robotics applications, evaluating perceptions systems based on these standard metrics is desirable but insufficient to comprehensively characterize system performance. The dimension missed is the robot's ability to take action in ambiguous situations. Second, even if a vision algorithm had perfect performance on these off-line datasets, it is bound to encounter inputs that it cannot handle. During long-term autonomous operations, robotics systems have to contend with vast amounts of continually evolving, unstructured sensor data from which information needs to be extracted. In particular, the problem aggravates for perception systems where the problems are often ambiguous, and algorithms are designed to work under strong assumptions that may not be valid in real-world scenario. Several factors - degraded image quality; uneven illumination; poor texture - can affect the quality of visual input, and consequentially affect the robots' action, possibly resulting in catastrophic failure. For example, over- or under-exposed images, images with massive motion blur would confound a vision-based guidance system for a mobile robot. While it is possible to anticipate some of these conditions and

S. Daftry, S. Zeng, J. A. Bagnell and M. Hebert are all members of the Robotics Institute, Carnegie Mellon University, Pittsburgh, PA, 15213, USA. This work was supported by the ONR MURI grant for 'Provably-Stable Vision-based Control of High-speed Flights through Forest and Urban Environments'. Email: {daftry, slzeng, dbagnell, hebert}@cmu.edu

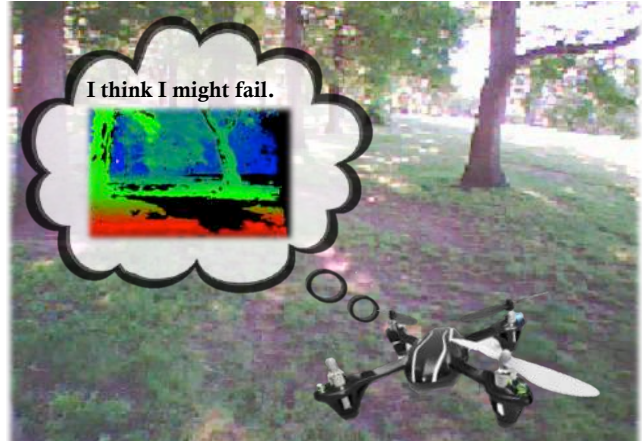

Fig. 1. Robot Introspection: Having a self-evaluating facility to *know when it doesn't know* will augment a robot's ability to take reliable mission-critical decisions in ambiguous situations.

hardcode ad-hoc test to reject such inputs, it is impossible to anticipate all of the conditions encountered in a real world environment.

This presents a challenge as the real cost of failure can be significant. For example, an autonomous drone that fails to detect an obstacle can suffer disastrous consequences. They can cause catastrophic physical damage to the robot and its surroundings. We thus believe that it is essential to identify as early as possible, i.e., based on sensor input, situation in which a trustable decision cannot be reached because of degraded performance of the perception system. Crucially, and central to this paper, this requires the perception system to have the ability to evaluate its input based on a learned performance model of its internal algorithm. We call this capability *introspection*[1], a precursor to action selection.

In this paper, we explore the possibility of building introspective perception systems that can reliably predict their own failures and take remedial actions. We argue that while minimizing failures has been the primary focus of the community, embracing and effectively dealing with failures has been neglected. We propose a generic framework for introspective behavior in perception systems that enables a robot to *know when it doesn't know* by measuring as to how qualified the state of the system it is to make a decision. This helps the system to mitigate the consequences of its failures by learning to predict the possibility of one in advance and take preventive measures.

---

[1]**Introspection.** The act or process of self-awareness; contemplation of one's own thoughts or feelings, and in the case of a robot, its current state.

We explore these ideas in the context of autonomous flight, where mission-critical decisions equate to safety-critical consequences [3]. Consider the task of autonomously navigating a dense cluttered environment with a MAV using monocular vision. This requires the MAV to successfully estimate a local map of its surrounding in real-time, at minimum the depth of objects relative to the drone, and based on this map to select an optical trajectory to reach its destination, while avoiding any obstacle. It is crucial to note that this kind of a system, like most other real-world robotics applications, involves modular pipelines, where the output of one module is fed into another as input. In such cases, decisions based on overconfident output from one subsystem, in this case a perception module, will lead to catastrophic failure. In contrast, anticipation of a possible failure based on previous experience allows for remedial action to be taken in the subsequent layers. Providing this more relevant prediction is the introspective quality that we seek.

The rest of the paper is organized as follows: In Section II, we review the related work on introspection. In Section III, we introduce our generic introspective perception framework. Section IV outlines our autonomous monocular navigation approach. Quantitative, qualitative and system performance evaluation with respect to related work is delineated in Section V.

## II. RELATED WORK

Our work addresses an issue that has received attention in various communities. The concept of introspection as introduced here can be closely related to active learning and incremental learning [4], [5], [6], where uncertainty estimates and model selection steps are used to guide data acquisition and selection. KWIK 'Knows What It Knows' frameworks [7] allow for active exploration which is beneficial in reinforcement learning problems. Further, reliably estimating the confidence of classifiers has received a lot of attention in the pattern recognition community [8]. Applications such as spam-filtering [9], natural language processing [10] and even computer vision [11] have used these ideas.

In robotics, the concept of introspection was first introduced by Morris et al. [12] and has recently been adopted more specifically for perception systems by Grimmett et al. [13], [14] and Triebel et al. [15]. They explore the introspective capacity of a classifier using a realistic assessment of the predictive variance. This is in contrast to other commonly used classification frameworks which often only rely on a one-shot (ML or MAP) solution. It is shown that the accountability of this introspection capacity is better for real world robotics applications where a realistic assessment of classification accuracy is required.

However, in all the above work, the confidence estimation is obtained by examining the output of the underlying system to assess its probability of failure on any given input. Thus, the evaluation methodology has to be designed specifically for each system. This is also particularly undesirable when the underlying system is computationally expensive. Instead, we explore the alternative approach of evaluating the input itself in order to assess the system's reliability. The main motivation for such an approach is that while the perception system may not be able to reliably predict the correct output for a certain input, predicting the difficulty or ambiguity of the input may still be feasible. Moreover, this is applicable to any vision system because the reliability estimate is agnostic to the actual algorithm itself and based on the input alone.

In a similar spirit, Jammalamadaka et al. [16] recently introduced evaluator algorithms to predict failures for human pose estimators. They used features specific to this application; Welinder et al. [17] and Aghazadeh and Carslsson [18] analyzing statistics of the training and test data to predict the global performance of a classifier on a corpus of test instances. Another line of research involves the domain of 'rejection' [19]. This is related to systems that identify an instance to belong to 'none of the above' class, as compared to the one's during training, and thereby refuse to make a decision on that instance. Zhang et al. [20] learn a SVM model to predict failures from visual input. Bansal et al. [21] learn a semantic characterization of failure modes.

The biometrics community has used image quality measures as predictors for matching performance [22]. The concept of input 'quality', as used here, is implicitly tied to the underlying perception system. Our work follows the same philosophy. Such methods, like ours, only analyze the input instance (e.g., fingerprint scan) to assess the expected quality of performance (e.g. fingerprint match). Our method is complementary to all the previous approaches: it uses generic image appearance features, provides an instance-specific reliability measure, and the domain we consider is not only concerned with unfamiliar instances that arise from discrepancies between the training and test data distributions but also addresses familiar but difficult instances. In addition, we show the effectiveness of such predictions in a mission-critical decision making scenario such as robot navigation. We advocate that a similar level of self-evaluation should be part of all perception systems as they are involved in a variety of real-world applications.

## III. INTROSPECTIVE PERCEPTION

Introspective robot perception is the ability to predict and respond to unreliable operation of the perception module. In this section, we describe a generic framework for introspective robot perception, which we illustrate in the context of vision-based control of MAVs in the next section.

### A. Introspection Paradigm

Introspection is distinct from conventional elements of robot architecture in that introspective processes do not reside in the path of architectural data flow [12]. Instead, it envelops all other layers. This allows the introspective system an end-to-end observation of the robot's computational activity. Through observation, introspection then produces estimates of the reliability of current state and delivers system-level feedback to the deliberative facilities.

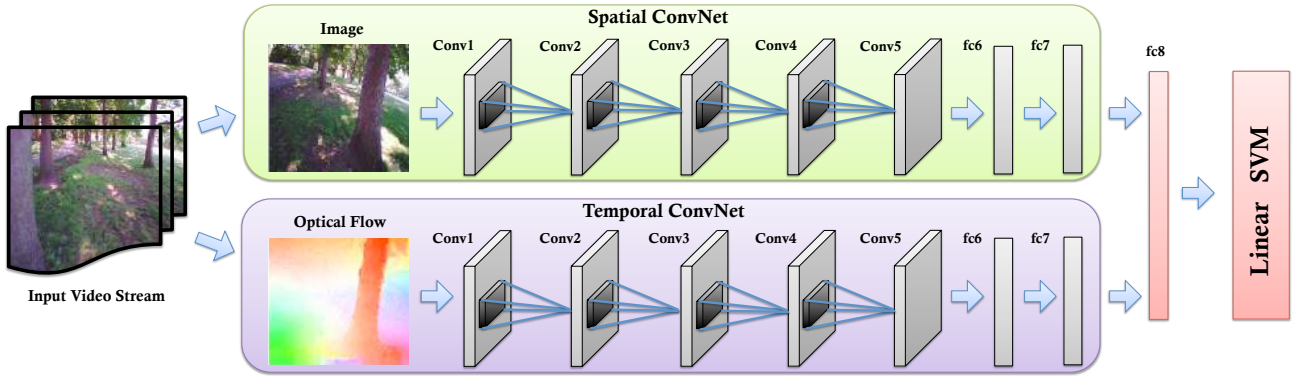

Fig. 2. Spatio-temporal convolutional neural net architecture for deep introspective perception. The layer schematics are same as AlexNet [23]

The problem statement can thus be formulated as: To learn a model for the performance of a perception system relative to a specific task, e.g., trajectory evaluation for autonomous flying, solely based of the visual input; and use it to reliably predict whether the system is likely to fail on a given test instance. We frame this as a multi-step supervised learning algorithm: A deep spatio-temporal convolutional net is trained to learn good invariant hidden latent representations. The resulting features are then used as input to train a linear SVM [24], which generates a failure prediction score as output. SVMs in combination with CNN features are a widely used alternative to softmax for classification and regression tasks [25]. The output score is between 0 and 1 with higher scores indicating images for which the perception algorithm is predicted to be unreliable. The score can then be used by the rest of the system to decide whether to plan an action based on this input or to discard it and generate an alternate behavior.

### B. Learning Spatio-Temporal ConvNets

In recent years, deep convolutional neural net (CNN) based models and features [23] have been proven to be more competitive than the traditional methods on solving complex learning problems in various vision tasks [26]. Now, while great progress have been made in the domain of single images for multimedia applications, the temporal component in the context of videos is generally more difficult to use. We argue that in robotic applications video data is readily available, and can provide an additional (and important) cue for any perception task.

We advocate that the use of this temporal component is crucial to introspective perception. The challenge, however, is to capture complementary information from still frames and motion between frames. To that end, we propose a two-stack CNN architecture as shown in Figure 2, which incorporates spatial and temporal networks. This architecture is reminiscent of the human visual system processes which uses the ventral pathway for processing the spatial information [27], such as shape and color, while the dorsal pathway is responsible for processing the motion information [28].

The layer configuration of our spatial and temporal ConvNet is similar to AlexNet [23]. Both the spatial and temporal stacks contain five feature extraction layers of convolution and max-pooling, followed by two fully connected layers. Dropout is applied to the two fully connected layers. The spatial stack operates on individual video frames, while the input to our temporal ConvNet stack is formed by trajectory stacking [29] of optical flow displacement fields between several consecutive frames. Simonyan *et al.* [30] have shown that such an input configuration, in the context of action recognition from videos, explicitly describes the motion between video frames. This makes the learning easier, as the network does not need to estimate motion implicitly. The outputs of the fc7 layers from both the stacks are concatenated and fed to the fc8 layer, which builds a shared, high dimensional hidden representation.

The output of fc8 layer goes to the softmax loss layer which estimates the deviation from the given target. This loss is then used to compute the gradients needed for back-propagation. During inference time, the loss layer is not used anymore and the output of fc8, represented as a d-dimensional feature vector: $x_i \in \mathcal{R}^d$, constitutes the deep features we seek. Responses from the higher-level layers of CNN have proven to be effective generic features with state-of-the-art performance on various image datasets [26]. Thus, we use these responses as generic features for introspection.

### C. Quantifying Introspection

In order to characterize the introspective capacity of a perception framework, a quantifiable measure is required. During training, we assume that we are given a dataset of $N$ instances, $\mathcal{X} = \{x_1, ..., x_N\}$ and targets $\mathcal{Y} = \{y_1, ..., y_N\}$. The targets $y_i$ are the task-dependent performance or accuracy score for the inputs $x_i$. This may be any metric that is representative of the failure probability of the system. For example, if the underlying perception system is a binary classifier, $y_i$ can be the 0-1 loss. If the task is semantic segmentation, we can use proportion of correctly classified pixels as a measure of accuracy. SVM training then involves learning a set of parameters for regression task.

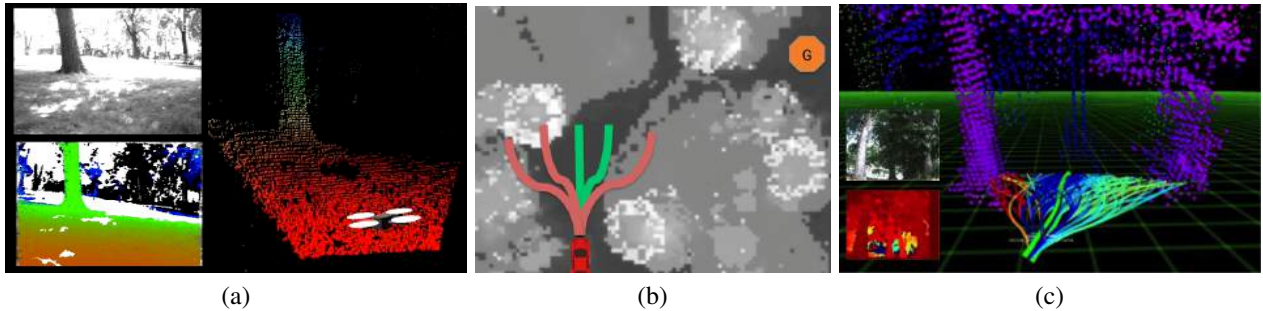

Fig. 3. (a) Perception module: example image, depth map and a bird's eye view of the local 3D scene map. (b) Illustration of receding horizon control (c) Receding horizon control using depth images from stereo. Red trajectories indicate that they are more likely to be in collision. Note: Best seen in color.

## IV. VISION BASED AUTONOMOUS NAVIGATION

In this section we describe our vision-based autonomous navigation framework in the context of autonomous flight through a dense, cluttered forest. This application is particularly interesting given the wide spectrum of challenges involved in fault tolerance control in perception for autonomous aerial systems navigating outdoor diverse environments based on a single sensor [32].

### A. Setup

We use a modified version of the 3DR ArduCopter with an onboard quad-core ARM processor and a Microstrain 3DM-GX3-25 IMU. Onboard, there are two monocular cameras: one PlayStation Eye camera facing downward for real time pose estimation and one high-dynamic range PointGrey Chameleon camera for monocular navigation. The image stream from the front facing camera is streamed to the base station, where the perception and planning module is processed. The optimal trajectory is transmitted back to the onboard computer where the control module does trajectory tracking.

### B. Perception

The perception system runs on the base station and is responsible for providing a semi-dense 3-D scene structure which can be used for motion planning. Our method is based on the recent progress in direct visual odometry based approaches [33], [34]. In the last few years, such direct approaches for scene geometry reconstruction have become increasingly popular. Instead of operating solely on visual features, these methods directly work on the image intensities for both mapping and tracking: The world is modeled as a dense surface while in turn new frames are tracked using whole-image alignment.

**Mapping:** The depth measurements are obtained by a probabilistic approach for adaptive-baseline stereo [33] and converted to an inverse-depth representation. The inverse depth map is propagated to subsequent frames, once the pose of the following frames have been determined and refined with new stereo depth measurements. Based on the inverse depth estimate $d_0$ for the pixel, the corresponding 3D point is calculated and projected into the new frame and assigned to the closest integer pixel position, providing the new inverse depth estimate $d_1$. An example of the obtained depth estimates, and the corresponding map is shown in Figure 3a.

**Tracking:** Given an image $I_M : \Omega \to \mathbf{R}$, we represent the inverse depth map as $D_M : \Omega_D \to \mathbf{R}^+$, where $\Omega_D$ contains all pixels which have a valid depth. The camera pose of the new frame is estimated using direct image alignment. The relative pose $\xi \in SE(3)$ of a new frame $I$, is obtained by directly minimizing the photometric error:

$$E(\xi) := \sum_{x \in \Omega_{D_M}} \|I_M(x) - I(w(x, D_m(x), \xi))\|_\delta$$

, where $w : \Omega_{D_M} \times \mathbf{R} \times SE(3) \to \omega$ projects a point $x$ from the reference frame image into the new frame and $\|\cdot\|_\delta$ is the Huber norm to account for outliers. The minimum is computed using iteratively re-weighted Levenberg-Marquardt minimization [35].

### C. Planning and Control

We use a semi-global planning approach in a receding horizon control scheme [36]. Once the planner module receives a depth map from the perception module, the local point cloud is updated using the current pose of the MAV. A trajectory library of 78 trajectories of length 5 meters is budgeted and picked from a much larger library of 2401 trajectories using the maximum dispersion algorithm by Green et al. [37]. Further, for each of the budgeted trajectories a score value for every point in the point cloud is calculated by taking into account several factors, including the distance to goal, cost of collision, etc., and the optimal trajectory to follow is selected. The control module takes as input the selected trajectory to follow and generates waypoints to track using a pure pursuit strategy [38]. Using the pose estimates of the vehicle, the MAV moves towards the next waypoint using a generic PD controller. For further details on planning, we direct the reader to our previous work [39].

### D. Introspection

To emulate the kind of introspective behavior we are looking for in navigation tasks, we propose to learn an introspection model, as described in Section III, to reliably predict when an image is going to give a poor estimate of the trajectory labels. Hence, to train the introspection model for this application, the task-based performance score $y_i$ is defined as the fraction of trajectories that are correctly predicted as collision-free or collision-prone, with respect to ground truth. Figure 3b illustrates an example of receding horizon control on an autonomous drone.

A pre-selected set of dynamically feasible trajectories of fixed length (the horizon), is evaluated on a cost map of the environment around the vehicle and the trajectory and the most optimal trajectory is selected. The challenge here, however, is to obtain ground-truth cost map, as it would require manual annotation of each frame over several hours of video. Rather, we use a self-supervised method to obtain ground truth by obtaining corresponding 3D map using stereo data as shown in Figure 3c. We collected a dataset - BIRD Dataset, as detailed in Section V for training and validation of our introspection model. During data collection flights, we run two pipelines in parallel: one using our described perception system, and one with the registered stereo depth images. Trajectory data from both pipelines are recorded simultaneously while flying the MAV, and we treat the one from stereo as the ground truth.

**Emergency Maneuvers:** During autonomous flight, inference is done on the learned model for each input frame to produce a failure probability score. If the score is above a given threshold (0.5 for all our experiments), the system is directed to execute a set of emergency maneuvers. Apart from preventing the drone from an inevitable crash, these emergency maneuvers need to additionally ensure that they help make the system more reliable by taking images under different conditions, than the ones under which the frame generating the failure was taken. Thus, the design of these maneuvers needs to take into account the failure modes specific to the system. A visual perception system, like ours, is strongly susceptible to sudden motions like pure rotation and image exposure. Thus, when *alerted*, a set of pure translational trajectories constrained in the image plane pointing in different directions to avoid exposure issues are executed for $N = 3$ seconds till the reliability is improved, after which normal deliberative planning is resumed. Note: The set of emergency maneuvers used are hand designed to account for a few known failure modes. Generating these trajectories automatically is beyond the scope of this paper and has been left for future work.

## V. Experiments and Results

In this section we analyze the performance of our proposed method for introspective robot perception. All the experiments were conducted in a densely cluttered forest area as shown in Fig 4, while restraining the drone through a light-weight tether. It is to be noted that the tether is only for compliance to federal regulations and does not limit the feasibility of a free flight.

### A. BIRD Dataset

A Bumblebee color stereo camera pair ($1024 \times 768$ at $20$ fps) is rigidly mounted with respect to the monocular camera, and registered using an accurate calibration technique [40].

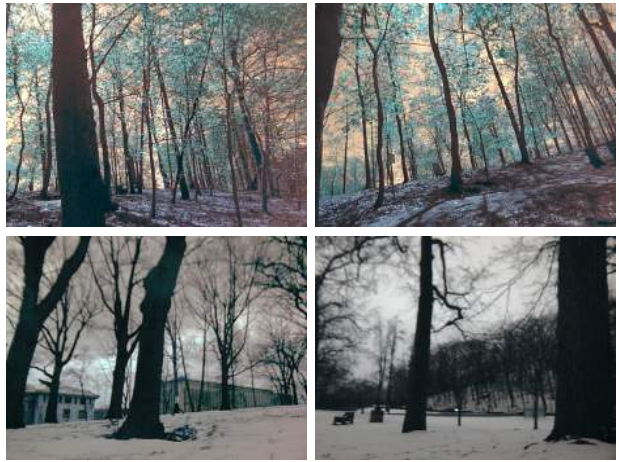

Fig. 4. BIRD dataset. Sample images depicting the diversity of data obtained over summer (Top) and winter (Bottom).

We collected data at several locations with varying tree density, under varying illumination conditions and in both summer and winter conditions. Our corpus of imagery with stereo depth information consists of $60\text{k}$ images. The training set comprises of $50\text{k}$ images, while $10\text{k}$ images have been used for validation. The training/test split has been done so as preserve contiguous sequences as opposed to random sampling from the corpus of images. While the latter will lead to training data which is better represented by test data, we believe that the former is perhaps more realistic for a robotics application.

We use the Caffe [41] toolbox and a NVIDIA Titan Z graphics card with $5760$ parallel GPU cores and $12$ gb of gpu ram for training our introspection model. The camera images are resized to $227 \times 227$ pixels for input to the spatial ConvNet. A multi-frame dense optical flow is computed for each frame using a total variational based approach [42], and fed to the temporal ConvNet. The convolutional layers (1-5) of the spatial network are pretrained on the ImageNet classification task [43]. Unlike the spatial stream ConvNet, which can be pre-trained on a large still image classification dataset like ImageNet, the temporal ConvNet needs to be trained on video data - and the BIRD dataset is still rather small. To decrease over-fitting, we use ideas from multi-task learning to learn a (flow) representation by combining several datasets in a principled way [44]. We use two datasets from action recognition: UCF-101 [45] with $13\text{K}$ videos and HMDB-51 [46] with $6.8\text{K}$ videos to pre-train the temporal ConvNet using the multi-task learning formulation.

### B. Quantitative Evaluation

**Risk-Averse Metric:** Quantitatively, we investigate the introspective capacity of the proposed model with its ability to trade-off the risk of making an incorrect decision with either taking remedial action or for the sake of this experiment, not making a decision at all. We believe such a metric is crucial when dealing with safety-critical applications and thus, present the Risk-Averse Metric (RAM). This metric was first proposed by Zhang *et al*. [20] and we adopt it to describes an Error vs Failure Rate (EFR) curve. Failure Rate is defined as the proportion of test images on which the perception system does not output a decision, in fear of providing an incorrect decision. The curve is computed by sorting the test images in descending order of their reliability as estimated by inference on our trained model. We then retain only a FR proportion of the test images (FR [0, 1]), and discard the rest. Since, the output of our perception system is a local 3D map, we compute the accuracy measure as mean reconstruction error with respect to stereo ground truth on these retained images and plot error vs. FR. For other applications like classification or segmentation, it would be more natural to use accuracy instead of error. If introspection were perfect, it would discard the worst performing images. Error would be low at low FR and then increase gracefully as FR tends to 1. If introspection performed at chance level, on average, the accuracy would remain constant with varying FR. We compare the performance of our method to these upper and lower bounds (Figure 5a). We see that an approach even as straightforward as introspection can perform significantly better than chance.

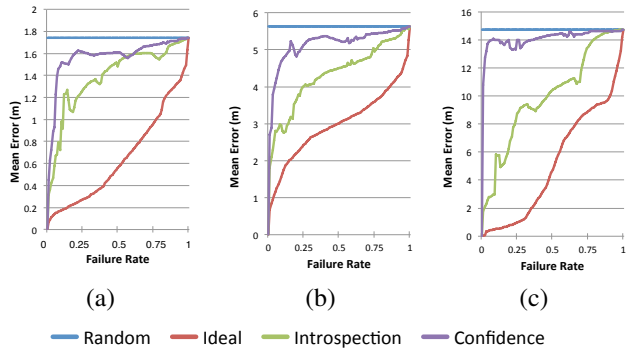
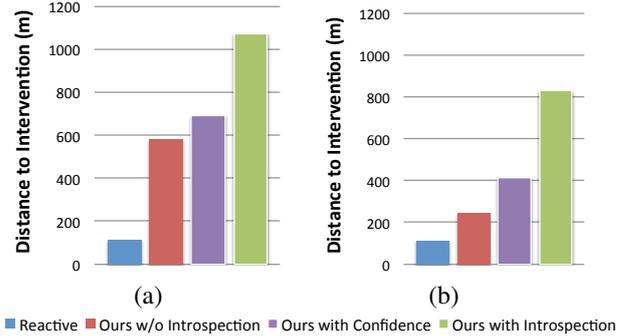

Fig. 5. Error vs. Failure Rate curves for depth estimation using (a) Direct VO (b) Depth CNN and (c) Non-linear Regression

Fig. 6. Average flight distance per intervention for (a) Low Density and (b) High Density regions.

**Performance Benchmarking:** Many vision systems produce indicators of the confidence of their *output* e.g. energy of a conditional random field for the task of semantic segmentation or the uncertainty of a classifier [14]. As discussed in Section II, previous methods have used this confidence score as an estimate for failure probability. For our task, we treat the mean per-pixel variance for depth estimates as an estimate of uncertainty. Thus, during inference, we classify an image as failure if the uncertainty is higher than a given threshold. As baseline, we compare and empirically show an advantage over using such system-specific failure detection approach. We plot the results for RAM in Fig 5 along side that from our introspection model. While model-based confidence estimation can thought to be more reliable since it knows the inner workings of the underlying algorithm, it is interesting to note that a black-box approach, like ours, still performs better. This can be explained by the fact that analyzing specific error modes is not able to capture the entire spectrum of possible failures. For systems that can afford to evaluate output confidence, combining it with our approach would yield a better introspection model.

**System-Agnostic Performance:** In order to show that our introspection model is generic and agnostic to the underlying perception system, we performed similar experiments with two other diverse approaches for depth map estimation - Depth CNN [47] and Non-linear Depth Regression [39] as can be see in Figure 5 (b-c). The results show that indeed the introspection model captures information orthogonal to the underlying perception system's beliefs. Moreover, it can be observed that for an unreliable system (one with higher mean error) such as non-linear depth regression, the performance of our introspection model in risk aversion is far superior to that of a confidence based estimator which performs only close to chance due to its dependence on the unreliable system itself.

**Computational Advantage:** Our approach has an implicit computational advantage. It needs to run the underlying vision system to obtain the targets $y_i$ only during training. During inference, we estimate the reliability of the system using the raw input image stream only. This makes the introspection model computationally better as any confidence based method would first need to run the base system to obtain its output. This also makes our system preemptive - allowing early rejection of possible failures.

### C. Introspection Benefits to Autonomous Navigation

In this section, we study the benefits of our proposed introspection model to the application of autonomous navigation. Quantitatively, we evaluate the performance of the obstacle avoidance system by observing the average distance flown autonomously by the MAV over several runs (at 1.5 m/s), before an intervention. An intervention, in this context, is defined as the point at which the pilot needs to overwrite the commands generated by our control system so as to prevent the drone from an inevitable crash. Experiments were performed with and without using our proposed introspective perception approach. Results comparing to previous work on monocular reactive control by Ross *et al.* [48] as baseline has been shown in Figure 6. Tests were performed in regions of low and high clutter density (approx. 1 tree per $6 \times 6$ $m^2$ and $12 \times 12$ $m^2$, respectively). It can be observed that introspective navigation results in significantly better performance. In particular, the drone was able to fly autonomously without crashing over a 1000 m distance at average. The difference is higher in case of high-density regions where committing to an overconfident decision can be even more fatal.

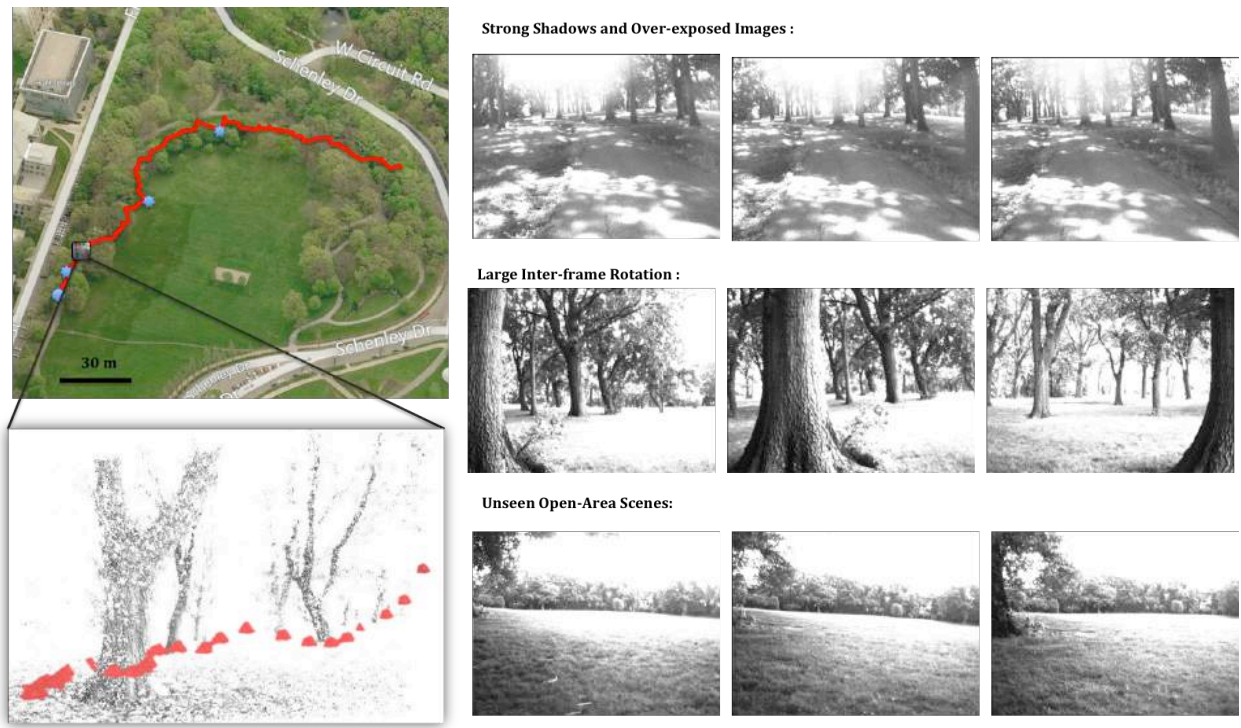

Fig. 7. Qualitative results of our proposed introspective perception approach for autonomous navigation. (Top Left) Testing area near Carnegie Mellon University, Pittsburgh, USA. The flight path for one of the test runs has been overlayed on an aerial image of the test site. The instances where the system was alerted have been marked in blue asterisk. (b) A region of the path has been zoomed to show the quality of 3D map reconstructed and the corresponding flight path within the dense forest environment. (c) Row 1-3 illustrates the stream of images when system was alerted, along with the intuitive reasoning.

### D. Qualitative Evaluation

Further, we extend our evaluation to qualitatively assess our introspection model during flight tests. In Figure 7a, we show the flight path taken by the MAV during a test run. The points at which the introspection model *alerted* the system of a possible failure has been marked with an asterisk in blue. Now, neural networks are known to largely remain 'black-boxes' whose function is hard to understand. Additionally, since we do not have any ground-truth during system flight tests, it is very hard to semantically characterize predicted failures into failure modes. However, we try to analyze the predicted failures, with the aim to obtain some intuition towards the kind of failures our model is learning. Figure 7c shows the stream of images predicted to be unreliable. Each row shows a different scenario of failure that occurred during the run. The model correctly predicted failures for set of images that are corrupted with strong illumination effects like over-exposure (Row-1); the effectiveness of the temporal model is evident as the strong inter-frame rotations are flagged (Row-2); and finally, previously unseen scenarios like open grounds are again flagged as uncertain (Row-3). We only present the most interesting results here; See attached video for more qualitative results and observe introspective perception in action.

## VI. CONCLUSION

In this paper, we introduced the concept of introspective perception - a generic system-agnostic framework for learning how to predict failures in vision systems. We advocate that every perception system needs to be equipped with such an introspective check in order to be self-aware and allow for remedial actions, when required. Moreover, we argue that introspection directly from raw sensor data is more effective than using uncertainty of model-based classifiers, and support this claim through qualitative and quantitative results. We have demonstrated our approach on a classical autonomous navigation task. However, our treatment and findings apply to any aspect of robotics where action is required based on inference from sensor data. Finally, with this work we hope to bridge the gap between the vision and robotics communities by taking a step in the direction of building self-evaluating vision systems that fail gracefully, making them more usable in real-world robotics applications even with their existing imperfections.